\documentclass[draft]{agujournal2019}
\usepackage{url} 
\usepackage{lineno}
\usepackage[inline]{trackchanges} 
\usepackage{soul}

\draftfalse

\journalname{Journal of Advances in Modeling Earth Systems (JAMES)}

\begin{document}

%
%


\title{Using Explainability to Inform Statistical Downscaling Based on Deep Learning Beyond Standard Validation Approaches}

%
%




\authors{Jose Gonz\'alez-Abad\affil{1}, Jorge Ba\~no-Medina\affil{1}, Jos\'e Manuel Guti\'errez\affil{1}}


\affiliation{1}{Instituto de F\'{\i}sica de Cantabria (IFCA). CSIC - Universidad de Cantabria, Santander, Spain}




\correspondingauthor{Jose Gonz\'alez-Abad}{gonzabad@ifca.unican.es}




\begin{keypoints}
\item Deep learning has shown promise for downscaling, but still lacks confidence for climate change due to lack of explainability

\item Explainable artificial intelligence (XAI) facilitates the evaluation of deep downscaling models by unravelling their internal behaviour

\item XAI techniques can detect structural problems that are not revealed by standard evaluation methods
\end{keypoints}

%
%

%
%


\begin{abstract}
Deep learning (DL) has emerged as a promising tool to downscale climate projections at regional-to-local scales from large-scale atmospheric fields following the perfect-prognosis (PP) approach. Given their complexity, it is crucial to properly evaluate these methods, especially when applied to changing climatic conditions where the ability to extrapolate/generalise is key. In this work, we intercompare several DL models extracted from the literature for the same challenging use-case (downscaling temperature in the CORDEX North America domain) and expand standard evaluation methods building on eXplainable artifical intelligence (XAI) techniques. We show how these techniques can be used to unravel the internal behaviour of these models, providing new evaluation dimensions and aiding in their diagnostic and design. These results show the usefulness of incorporating XAI techniques into statistical downscaling evaluation frameworks, especially when working with large regions and/or under climate change conditions.
\end{abstract}

\section*{Plain Language Summary}
Due to limitations in the computational resources available, General Circulation Models (GCMs) are advocated to simulate the climate system over coarse resolution grids. This hampers the applicability of GCM products in the regional-to-local scale, highly demanded by different socio-economic sectors. Statistical downscaling aims to solve this problem by generating high-resolution climate fields. Recently, machine learning techniques (particularly deep learning models) have shown promising results in this task. These models are first trained in a historical period through observational datasets, and then applied to the GCM outputs of plausible far-future scenarios, thus generating high-resolution climate change products. To assess the plausibility of the derived downscaled fields, several validation frameworks are performed, (e.g., skill to reproduce the present climate) which aim to assess the generalization of the models. Here, we present a novel evaluation protocol building on eXplainable Artificial Intelligence (XAI) to examine the suitability of certain deep learning models for climate downscaling. We frame the analysis on a use-case aiming to downscale temperature over North America. Contrary to standard validation approaches where the intercomparison results show no remarkable differences across models, we find the XAI protocol to be useful to diagnose failures relevant when applying these models to far-future scenarios. 

%
%

%


%
%
%
%

\section{Introduction}

Global Climate Models (GCMs) simulate the spatio-temporal evolution of the climate system forced by historical and future emission scenarios \cite{eyring_overview_2016}. However, computational and physical limitations constrain the spatial resolution of the GCM  outputs (100-200 km for state-of-the-art simulations) and misrepresent local processes \cite{maraun_statistical_2018}. These limitations hamper the applicability of GCMs at the regional scale, where climate information is highly demanded to conduct impact studies and elaborate adaptation plans for different socio-economic sectors.

Empirical Statistical Downscaling (ESD) bridges this gap by learning a model/link between a set of large-scale atmospheric fields (predictors) and the regional/local variable of interest (predictand) \cite{maraun_statistical_2018}. Under the Perfect Prognosis approach (PP), statistical models are trained using simultaneous predictor-predictand observational records (typically reanalysis and gridded/point observations) over a 20-30 years period characterizing recent climate. The resulting model is then fed with the predictors from the GCM simulations (e.g. historical and future projections) to produce downscaled information.

Deep convolutional neural networks (CNNs, \cite{Goodfellow_deep_2016,lecun_convolutional_1995}) have been recently introduced as a promising PP downscaling technique due to their ability to automatically infer spatial features that encode predictive information from predictor fields.
Different studies have examined the performance of these methods for downscaling in both present \cite{pan_improving_2019, bano_configuration_2020, sun_statistical_2021} and future climate conditions \cite{bano_suitability_2021, bano_downscaling_2022}, with successful results. Moreover, these advances have led to the production of continental-wide projections following the CORDEX protocol constituting the first continental-wide CORDEX contribution with PP statistical downscaling methods \cite{bano_downscaling_2022}. However, the complexity and black-box nature of CNNs raise some concerns about their widespread application to generate actionable downscaled results. In particular, existing evaluation methodologies do not address these concerns, as they do not help to diagnose and explain potential problems associated with the choice of models (different components and designs), such as structural deficiencies or their ability to extrapolate/generalise.

The field of eXplainable Artificial Intelligence (XAI) \cite{zhang_survey_2021,buhrmester_analysis_2021,das_opportunities_2020} has recently emerged to address the growing need to explain the functioning and results of deep learning models, making their internal structure understandable in human terms. The application of these techniques in the downscaling context is still incipient, with some promising results \cite{bano_understanding_2020,rampal_high_2022} proving that CNNs are able to identify the relevant predictors and the spatial regions of influence, thus performing an automatic predictor selection in their hidden layers. 

In this paper we use XAI techniques for the evaluation of downscaling methods, enhancing and complementing the standard validation approach \cite{maraun_value_2015} with new diagnostics providing insight in the internal structure of the models and the importance of the different elements. We illustrate the proposed framework focusing on a challenging problem (downscaling temperature over the CORDEX North America domain) and conducting an intercomparison experiment using several DL methods proposed in the literature. We assess their performance in present and future climates beyond conventional validation metrics by applying new XAI-based metrics helping to identify structural problems which are not detected by standard validation methods.

The paper is structured as follows. Section \ref{sec2} introduces the data, the models intercompared and the XAI techniques used in this work. In section \ref{sec3} we describe the results of the experiments, focusing on the application of the XAI techniques. Finally in Section \ref{sec4}, the main conclusions of the work are presented.

\section{Experimental framework}\label{sec2}

\subsection{Region of study and data}\label{sec2:1}

\begin{figure}
    \centering
    \includegraphics[width=0.8 \linewidth]{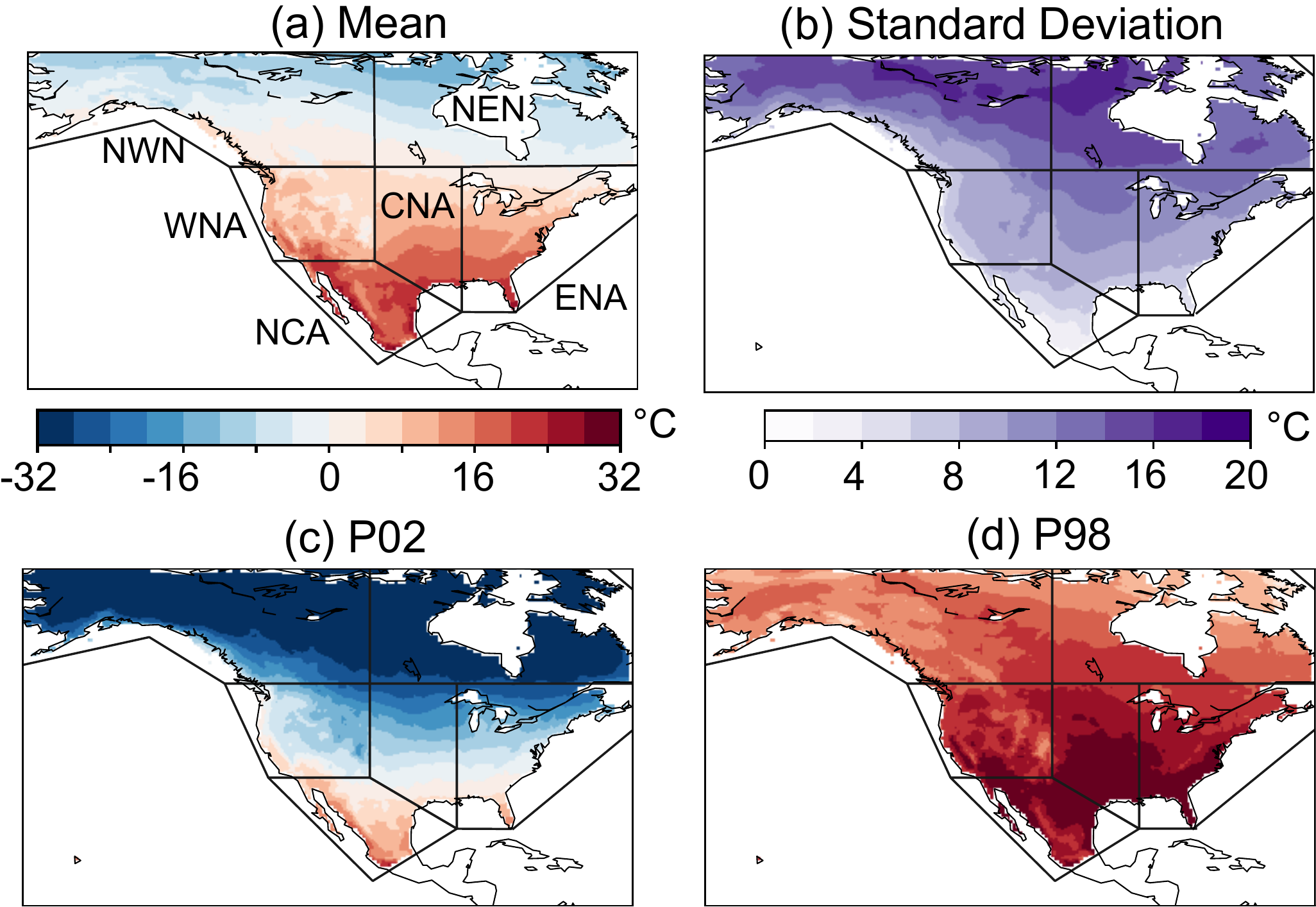}
    \caption{Climatology of the period 1980-2008 of EWEMBI near-surface air temperature for the mean, the standard deviation, and the 2nd (P02) and 98th (P98) percentiles. The panels display the five sub-continental IPCC reference regions over this domain, which are labelled in panel (a) as: Northwest North America (NWN), Northeast North America (NEN), West North America (WNA), Central North America (CNA), East North America (ENA) and North Central America (NCA), from left to right and top to down.}\label{fig:1}
\end{figure}

We focus on near-surface air temperature over the North America CORDEX domain (12$^\circ$-70$^\circ$ N and 60$^\circ$-165$^\circ$ W) using the regular gridded 0.5$^\circ$ EWEMBI daily dataset \cite{lange_earth2observe_2019} as the target predictand for statistical downscaling. 
Figure \ref{fig:1} displays the 2nd (P02) and 98th (P98) percentiles and the mean and standard deviation of EWEMBI's temperature for the region of study; it also shows the IPCC reference regions \cite{iturbide_update_2020} over the domain, which are used in this paper to provide regional results. This figures shows the complex climatology of the region with large latitudinal gradients for mean and extreme values and variability (e.g., from almost -15$^\circ$C in the north to 20$^\circ$C in the south for the mean). This makes downscaling a challenging problem over this continental-wide area, having to deal simultaneously with polar/tropical climates with high/low variability in the northern/southern areas. 

For the selection of predictors, we follow previous literature \cite{bano_configuration_2020, bano_suitability_2021} and use five large-scale variables (geopotential height, zonal and meridional wind, air temperature, and specific humidity) at four vertical levels (1000, 850, 700, and 500 hPa) at daily scale. The ERA-Interim reanalysis \cite{dee_era_2011} (regridded to a horizontal resolution of 2$^\circ$ using bilinear interpolation) is used together with the EWEMBI predictand for training, based on daily information for the common period 1980-2008.

The EC-Earth model run $r12i1p1$ \cite{doblas_using_2018} is used for testing the resulting downscaling models in historical (1980-2005) and future (2006-2100, from the Representative Concentration Pathway 8.5, RCP8.5, scenario) conditions. The EC-Earth model has been shown to properly reproduce key large-scale atmospheric patterns and consequently has been proposed for model assessment in several downscaling studies \cite{manzanas_dynamical_2018, bano_suitability_2021}. Moreover, the RCP8.5 scenario is selected since is the one describing the strongest climate change signal among those developed in the Coupled Model Intercomparison Project Phase 5 (CMIP5). As for ERA-Interim, we re-grid EC-Earth predictors to 2$^\circ$ from its native resolution (1.12$^\circ$) using bilinear interpolation. 

Taking into account the PP assumption that GCM predictors have to be realistically simulated (when compared with the ERA-Interim reanalysis in this case), following \cite{bano_downscaling_2022} we perform a signal-preserving adjustment of the monthly mean and variance of the GCM predictors to increase the distributional similarity with their counterpart reanalysis predictor fields. 

Finally, due to the sensitivity of deep learning algorithms to discrepancies in the scale of features, ERA-Interim and EC-Earth predictors are scaled. There are several ways to perform this scaling, in \cite{doury_regional_2022} authors rely on the daily spatial mean and standard deviation to scale predictors. To avoid the loss of temporal information due to this spatial scaling, they add as additional predictors the daily spatial mean and standard deviation time series of each predictor. Based on previous works \cite{bano_configuration_2020, bano_suitability_2021, bano_downscaling_2022} we follow a more common pre-processing scaling procedure in deep learning, the ERA-Interim and EC-Earth predictors are scaled at grid-box level using a single set of ---mean and variance--- parameters, thus not losing the temporal component. These parameters are computed from the training period of ERA-Interim data.

\subsection{Machine learning downscaling models}\label{sec2:2}

\begin{figure}
    \centering
    \includegraphics[width=\linewidth]{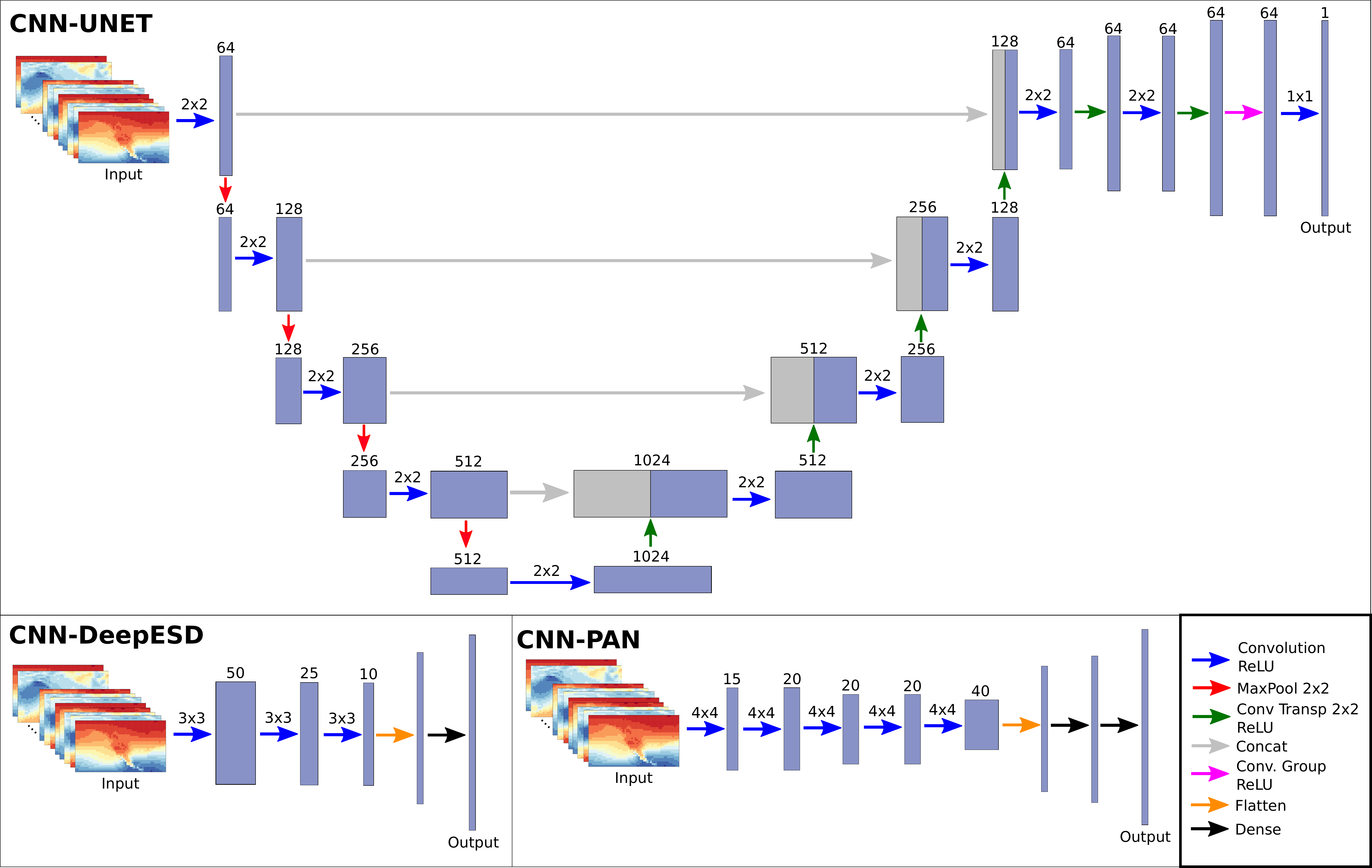}
    \caption{An overview of the topologies of the different CNN models intercompared in this study: CNN-UNET, CNN-DeepESD and CNN-PAN.}\label{fig:2}
\end{figure}

In this section we introduce the three CNN models used in this study representing the state-of-the-art of deep learning methods for downscaling (CNN-UNET, CNN-DeepESD and CNN-PAN, with increasing complexity in terms of number of parameters). CNN-UNET is fully convolutional consisting of several convolutional layers generating latent spatially consistent feature maps (or filter maps) from the predictor space optimized to explain the target variable; this model was originally proposed for image recognition tasks and subsequently adapted for climate downscaling as an out-of-the-shelf model. CNN-DeepESD and CNN-PAN combine a convolutional component (operating on the predictor input) with a subsequent dense component (linear in CNN-DeepESD and highly nonlinear in CNN-PAN), operating on the feature maps resulting from the convolutions to explain the target variables. These two models were specifically designed for climate applications, including statistical downscaling. Figure \ref{fig:2} provides an overview of the three models illustrating their different topologies (fully convolutional and convolutional plus dense).

All CNNs are trained following the same procedure, using ERA-Interim and EWEMBI as the predictor and predictand fields, respectively. The standardized variables are stacked as channels and passed to the input layer of the CNN models. The Mean Squared Error (MSE) is minimized using the ADAM optimizer \cite{kingma_adam_2014} with a learning rate of 1e-4 and a batch size of 100. We follow previous literature \cite{bano_configuration_2020} and split the observational data into a training and a test set. To prevent the models from overfitting we follow an early stopping strategy using a random $10\%$ split of the training data. 

\subsubsection{CNN-UNET}\label{sec2:2:1}
The UNET model is a popular CNN out-of-the-self configuration which has been widely used in image recognition problems \cite{ronneberger_unet_2015}. In the context of climate downscaling, UNETs have been used in different studies in Europe \cite{doury_regional_2022, quesada_repeatable_2022}. This model is composed of two different blocks: encoder and decoder (see Figure \ref{fig:2}). The first contracts the dimension of the input through convolutions and max pooling, while the second enlarges it to a specific dimension through transposed convolutions. Along the network, a path connecting the different levels of the encoder and decoder keeps them connected (see \cite{ronneberger_unet_2015} for more details).

We build a UNET model with the encoder composed of five blocks with a convolution, batch normalization and max pooling layer each --except the last block which does not have max pooling-- with 64, 128, 256, 512 and 1024 kernels respectively; the decoder is formed by a succession of four blocks with a transposed and standard convolutions with 512, 256, 128 and 64 kernels (see Figure \ref{fig:2} for details). All these layers use the rectified linear units (ReLU) activation function. After the decoder, a succession of standard and transposed convolutional layers are applied in order to get to the desired output dimensions. The absence of dense layers make this model fully-convolutional. 

\subsubsection{CNN-DeepESD}\label{sec2:2:2}
The CNN-DeepESD model was introduced in \cite{bano_configuration_2020} to downscale temperature and precipitation over the European domain, and was tested in climate change conditions in several studies over Europe \cite{bano_suitability_2021, bano_downscaling_2022}. CNN-DeepESD is composed of three convolutional layers with 50, 25 and 10 kernels with ReLU activation function for the hidden layers (see Figure \ref{fig:2} for details). The output of the last convolutional layer is flattened and passed to a dense layer with linear activation function, whose 10870 neurons represent the gridpoints of the predictand domain. Note that this is the simplest model combining convolutional and dense components (single linear layer in this case).  

\subsubsection{CNN-PAN}\label{sec2:2:3}
A CNN-PAN model combining convolutional and dense layers was introduced in \cite{pan_improving_2019} for downscaling daily precipitation at 14 different gridpoints representative of the climatology of North America. In this work we use the default network architecture proposed by the authors adapted to our task. The resulting model is similar to CNN-DeepESD but deeper (see Figure \ref{fig:2}), composed of five convolutional layers (with 15, 20, 20, 20 and 40 kernels with ReLU activation function) followed by two dense layers (with 5435 and 10870 hidden neurons, respectively, the latter representing the predictand gridpoints). As proposed in \cite{pan_improving_2019}, we explored max pooling, dropout and batch normalization layers with no successful results.

\subsection{eXplainable Artificial Intelligence (XAI)}\label{sec2:3}
Different XAI techniques have been designed to explain the functioning of deep learning models, making their internal structure understandable in human terms. In particular, saliency maps allow explainability by quantifying the influence of the input space features for a particular prediction. In the case of downscaling, saliency maps allow identifying the relevant predictor variables and the spatial regions of influence, thus facilitating diagnostic and explainability of the downscaled results. This plays a similar role as the feature importance algorithms used in standard machine learning techniques in previous statistical downscaling works \cite{he_spatial_2016,oh_machine_2022}.
 
Different saliency maps XAI techniques have been recently explored in the context of downscaling \cite{gonzalez-abad_interp_2022}, observing similar results across techniques, particularly among those relying on gradient computation. Following this, we apply a gradient-based technique known as Integrated Gradients (IG)  \cite{sundararajan_axiomatic_2017}, since it is known to overcome inherent problems of standard gradient-based methods ---e.g. gradient saturation \cite{glorot_understanding_2010}--- and has been used in other climate-CNN applications \cite{kondylatos_wildfire_2022}.

\begin{figure}
    \centering
    \includegraphics[width=\linewidth]{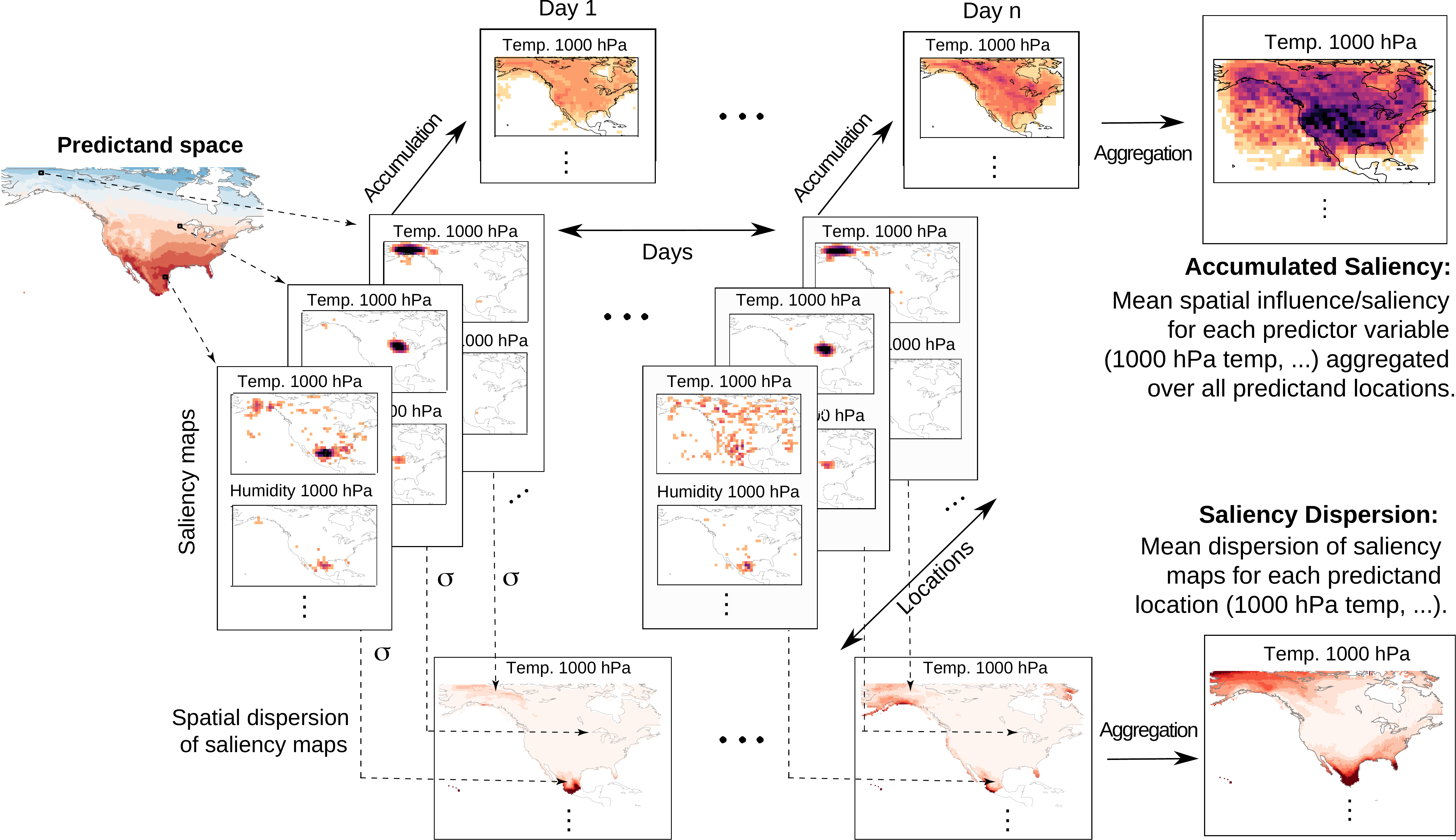}
    \caption{Schematic view of the XAI-based saliency metrics used in this work: Accumulated Saliency Maps (ASM) and Saliency Dispersion Maps (SDM). Saliency maps are computed for each particular output location (a predictand gridpoint) ---illustrated with three different north/central/south locations in the diagram---, for each observation (day) ---illustrated for two days, in columns, in the figure. Note that the ASM and the SDM fields have different dimensions, since the former is on the input (predictor) space and the latter on the output (predictand) one.}\label{fig:3}
\end{figure}

Saliency maps are computed by back-propagating gradients through the CNN layers, from the output layer back to the input features. These are computed for each neuron in the output layer ---corresponding to a particular predictand location/gridpoint--- for each observation (days in our case), as illustrated in Figure \ref{fig:3}. The result is a 3D saliency map defined on the predictor space (gridbox lon/lat - variable), where each value represents the contribution of a specific feature to the prediction of the model. Positive/negative salience values indicate inputs that contribute to increase/decrease the expected temperature at the target location.  Since we are only interested in the relevance of each input feature regardless of the sign, we take the absolute value. Following recent work \cite{toms_assessing_2021,mamalakis_investigating_2022} we normalize the saliency maps on daily basis (dividing by the maximum value independently for each predictand location), thus focusing on the identification of informative input variables and spatial regions and allowing the intercomparison of different models. Moreover, in order to reduce noise in the results avoiding the problem of gradient shattering \cite{mamalakis_investigating_2022}, we follow \cite{toms_assessing_2021} and remove relevance values below $0.1$.

In recent works, authors compute and study saliency maps for specific periods in order to unravel the behaviour of models for predicting particular (extreme) events \cite{rampal_high_2022}. However, similarly to feature importance techniques used in standard machine learning algorithms, in this work we are interested in the overall model behaviour over the whole (training/test/downscaling) climatic period. Therefore, we propose the following saliency magnitudes:

\begin{itemize}
\item \textbf{Accumulated Saliency Maps (ASM)}: Computed by accumulating the saliency maps corresponding to all the predictand locations thus obtaining the overall spatial influence/saliency for each predictor value for each particular gridbox (e.g. 1000 hPa temp, as represented at the top in Figure \ref{fig:3}). This is later aggregated within a specific period of time. This metric accounts for the overall importance of the different elements of the predictor space as contributing to the prediction of all the predictand locations. This is, the  ASM of a given predictor gridbox indicates the overall importance of this input to predict all the target location. High values indicate strong local influence and/or large influential area.

\item \textbf{Saliency Dispersion Maps (SDM)}: Saliency dispersion maps are computed by representing in each location of the (predictand) map the dispersion of the corresponding saliency maps; this value measures the spatial sparsity of the saliency map for a given target location, so the larger the dispersion the more scattered the influence of the predictor (e.g. 1000 hPa temp, as represented at the bottom in Figure \ref{fig:3}). As in the previous case, saliency dispersion maps are aggregated within a specific period of time. In particular, the dispersion of saliency maps can be calculated in several ways (entropy, standard deviation, etc.); in this work we use a weighted distance using as weight the geographical distance to the target location. Then these weighted values are summed, ending with a real value per location. To take into account the difference in area across the latitudinal axis, we rely on the Haversine formula \cite{robusto_cosine_1957} to compute the array of distances.
\end{itemize}

\section{Results}\label{sec3}
In this section we present the results obtained with the three CNN models intercompared in this study under different conditions. We first present the cross-validation results obtained under perfect conditions (i.e. with reanalysis predictors); then, we test their performance to extrapolate future climate conditions (using GCM predictors) and show that divergent results among the CNN models are obtained in some regions. Finally, we describe the application of the new XAI metrics introduced in this work that allow to partially explain these results.  The three CNN models are trained using ERA-Interim predictors and EWEMBI predictand as described in Sec. \ref{sec2:1}. 

\subsection{Standard evaluation: Cross-validation and extrapolation}\label{sec3:1}
\begin{figure}
    \centering
    \includegraphics[width=\linewidth]{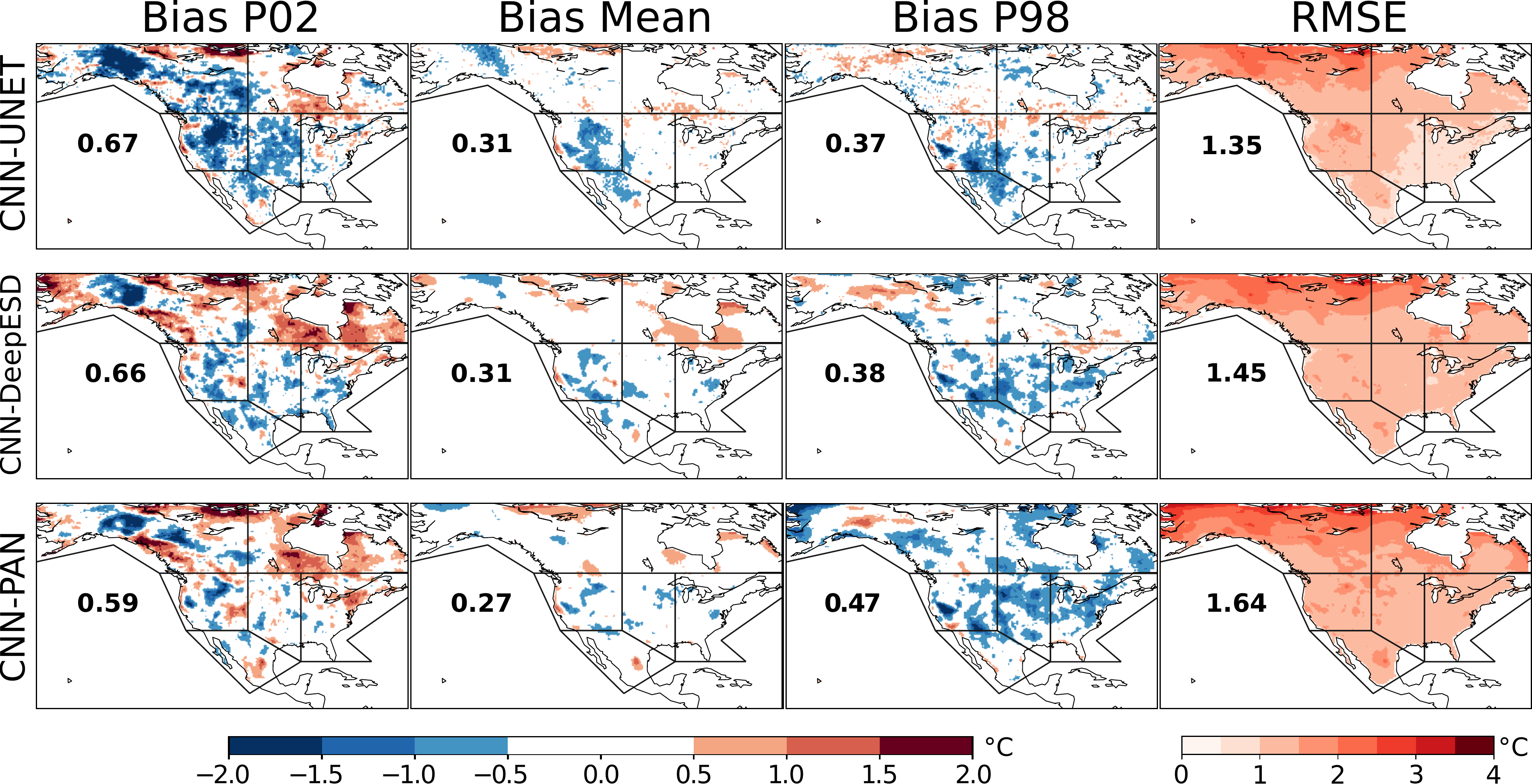}
    \caption{Validation metrics on the test set for the different CNN models. For each of these models (in rows) the biases of P02, mean and P98 are shown, together with the RMSE (in columns). The number in each panel represents the spatial mean of the absolute values. IPCC regions are also delimited.}\label{fig:4}
\end{figure}

Figure \ref{fig:4} displays the cross-validation results for four standard validation scores (biases of P02, mean, and P98, and Root Mean Square Error - RMSE) computed for the ERA-Interim test period to assess the overall performance of the models. The three CNN models are displayed in rows, in increasing order of complexity (according to the parameters involved), from the fully convolutional CNN-UNET to the convolutional and dense CNN-DeepESD and CNN-PAN models. For each of these validation metrics, the figure shows the spatial results as well as the spatial mean of the absolute values. The figure shows that the biases exhibit a similar spatial structure across models (particularly for CNN-DeepESD and CNN-PAN), with larger biases for the extremes (P02 and P98). CNN-UNET is the best performing model in terms of RMSE, particularly in the central eastern and southern regions. However, there is no best model outperforming the others in all regions for all scores and, overall, there is no reason to discard any of these models. 

\begin{figure}
    \centering
    \includegraphics[scale=0.3]{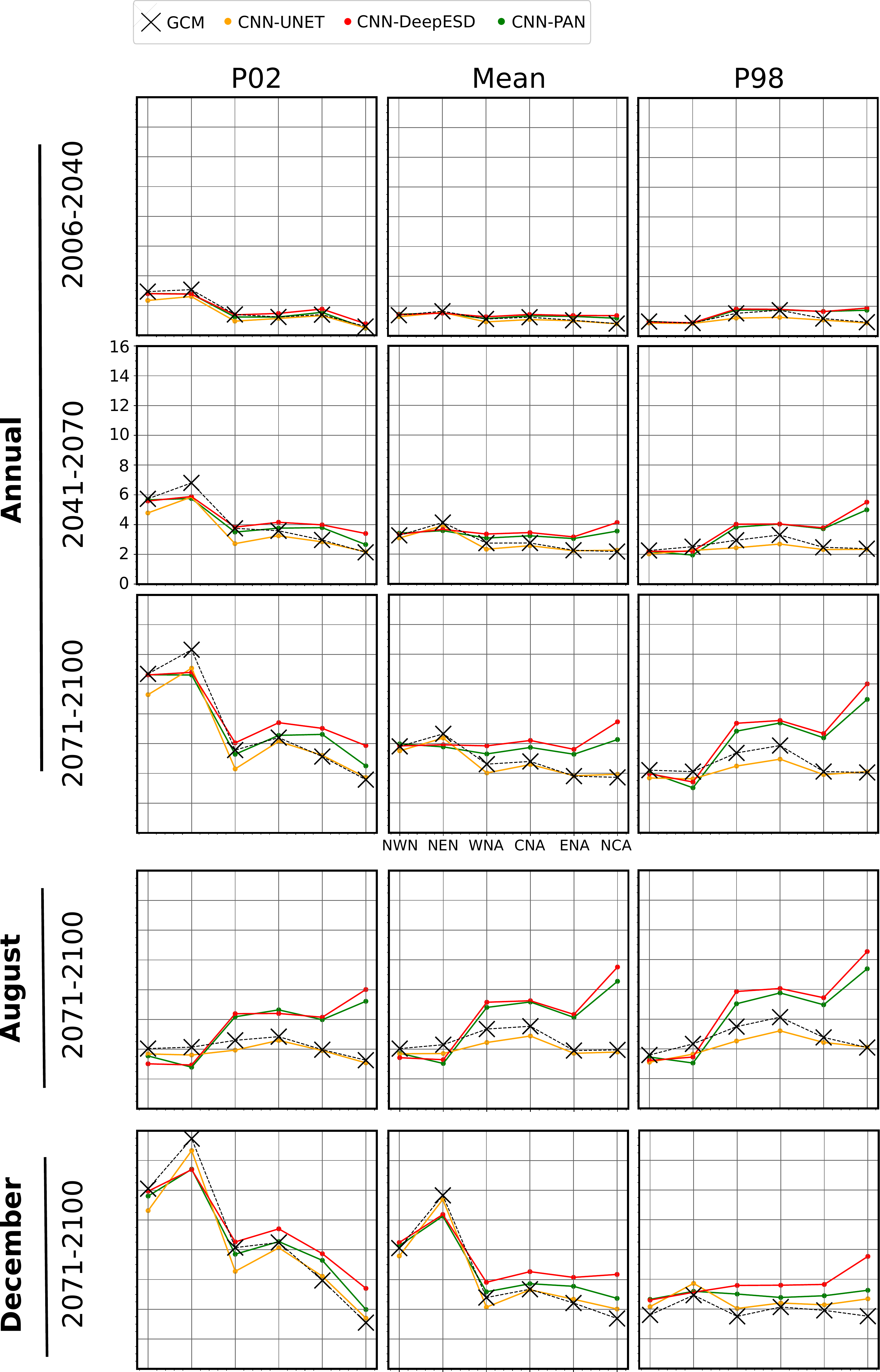}
    \caption{Delta change of the GCM and the CNN models for different future periods relative to the historical period (1980-2005), aggregated over the different IPCC regions (NWN, NEN, WNA, CNA, ENA, and NCA; see Figure \protect\ref{fig:1}), sorted from upper-left to bottom-right. The first three rows show the annual signals for the  periods 2006-2040, 2041-2070, and 2071-2100. The last two rows show the seasonal signals for the latter period, for August and December. For each of these periods the delta change is computed for the P02, mean and P98 (columns).}\label{fig:5}
\end{figure}

To test the extrapolation capability of the CNN models under future climate change conditions (when applied to predictors from GCM projections; see Section \ref{sec2:1}), we follow previous work and use the ``raw" GCM projections as pseudo-reality \cite{vrac_general_2007,bano_downscaling_2022}. We divide the future scenario into three different periods (2006-2040, 2041-2070 and 2071-2100) and compute the delta change between the future and historical (1980-2005) scenarios for the GCM and CNN models. Although downscaled projections may differ from the GCM signal (e.g. added value at local scale), the raw and downscaled climate change signals should be broadly consistent at the sub-continental scale (e.g. at the scale of the IPCC regions used in this paper). In Figure \ref{fig:5} we compare the delta change (climate change signal) across models for the different IPCC regions. Delta changes are computed for P02, mean and P98 indices (in columns) over the different intervals in the future (in rows). We also show the delta change on the last interval (2071-2100) for summer (August) and Winter (December) conditions. 

For the near-future period (2006-2040), the signals of the three CNN models seem to be broadly consistent with the signal of the GCM, except for the warm extremes (P98) where CNN-DeepESD and CNN-PAN slightly diverge from the GCM and CNN-UNET in the southern region (NCA). These differences increase progressively for mid- and future-term periods (2041-2070 and 2071-2100) in both warm (P98) and mean values for CNN-DeepESD and CNN-PAN, producing implausible results  in the central and Southern regions (WNA, CNA, ENA and NCA) with differences larger than 2 degrees in some cases. However, results for cold extremes (P02) are consistent with the GCM for all CNN models across the different periods.  The different performance of the CNN-DeepESD and CNN-PAN methods for cold and warm periods is further emphasized in the last two columns of Figure \ref{fig:5}  representing the results for cold (December) and warm (August) seasons.  Overall, the CNN-UNET model produced plausible results, with regional patterns similar to the GCM across all periods.  

The discrepancy between the CNN-UNET (producing plausible results when compared with the GCM) and CNN-DeepESD and CNN-PAN methods (with divergent climate change signals in the central and southern regions) is no apparent from the results obtained with the standard evaluation metrics presented in Figure \ref{fig:4}, where all models perform relatively well on the test set. These results highlight potential model deficiencies difficult to diagnose due to the complexity and black-box nature of the underlying models. In the following we see that the eXplainable Artificial Intelligence (XAI) techniques introduced in Sec. \ref{sec2:3} allow explaining these differences.

\subsection{Gaining explainability with XAI techniques}\label{sec3:2}
\begin{figure}
    \centering
    \includegraphics[scale=0.25]{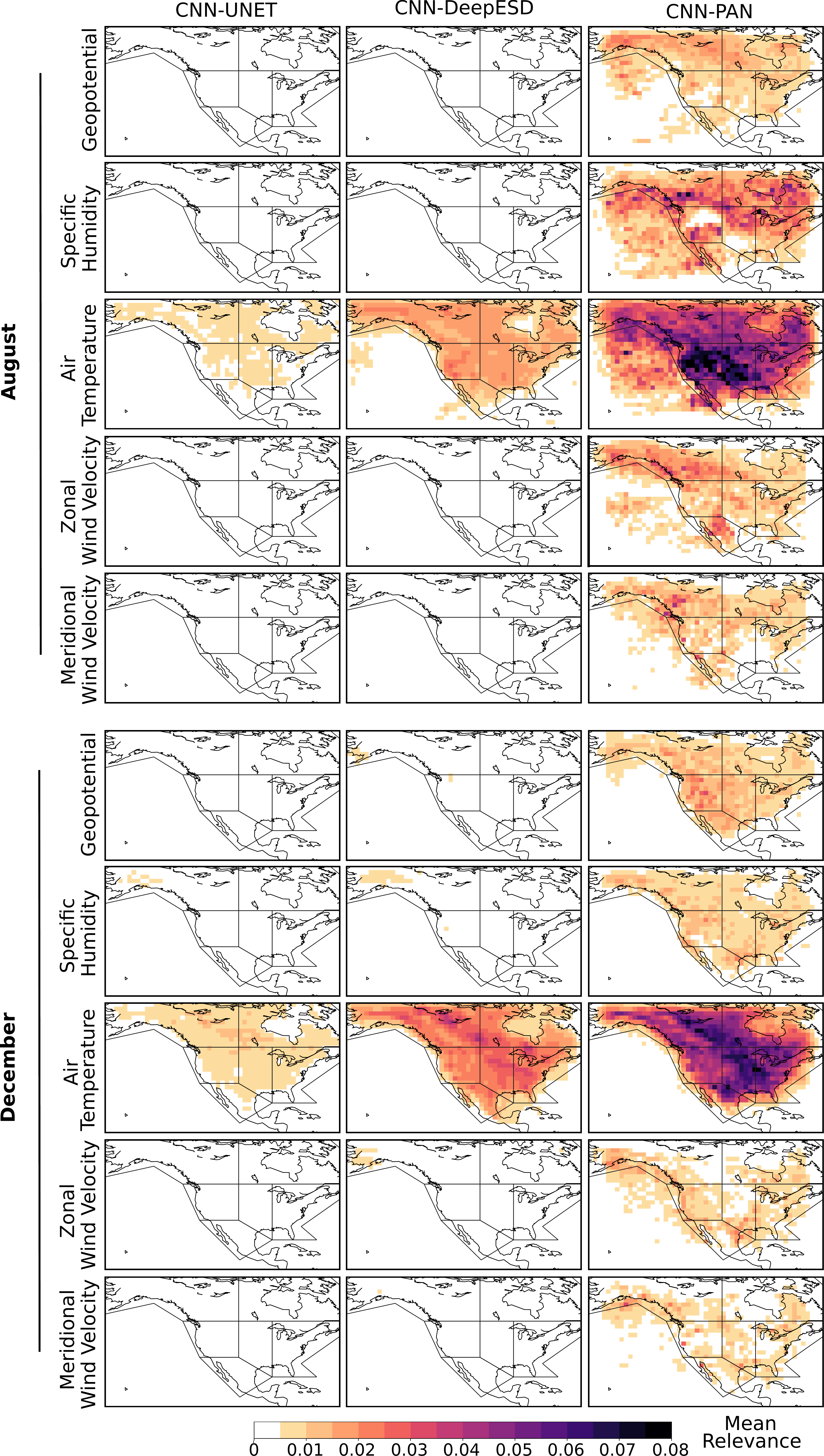}
    \caption{Accumulated Saliency Map (ASM) for August and December for each CNN model (in columns) for the five large-scale predictors (geopotential, specific humidity, temperature, and zonal and meridional wind velocity for a single 1000 hPa level).}\label{fig:6}
\end{figure}

The XAI explainability diagnostics introduced in Sec. \ref{sec2:3} provide  information to understand the performance of the CNN methods assessed in Sec. \ref{sec3:1}. In particular, the Accumulated Saliency Maps (ASM)  inform on the relevant predictors and regions of influence used by the models.  Figure \ref{fig:6} shows the August and December ASM aggregated over the test period.  For each CNN model (in columns), this figure shows the ASM corresponding to the five large-scale predictors at 1000 hPa (in rows); results for the other levels  (850, 700 and 500 hPa)  show lower saliency magnitudes and are not shown for simplicity. 

For the CNN-UNET and CNN-DeepESD models, temperature is the predictor that accumulates the highest relevance, although specific humidity also influences the prediction for December in the NWN region. The lower values shown by CNN-UNET indicate that the predictor's influence zones for the different target locations are more localised than in the case of CNN-DeepESD, which exhibits some weak influence areas even over the Pacific Ocean (for August). 

In the case of the CNN-PAN model, air temperature is also the most relevant variable, but the saliency values are also distributed among all the other predictors. We can see a noisier pattern with higher cumulative saliency over the whole of North America including also some large oceanic regions. This means that this model assigns relevance to predictor regions beyond the local areas of influence of the target locations.  This could be the result of the dense hidden layers included in this model, which act on the convolutions and capture wider areas of influence than the other models. The slightly worse test results obtained with this model could indicate a slight overfitting resulting from this additional complexity in the final layers. 

Note that the ASM provides an overall characterization of predictor importance over the whole domain, and allow comparison of the spatial extent of predictors' areas of influence in the different models, as indicated by the (spatially) cumulative saliency intensities. Saliency Dispersion Maps (SDM) provides complementary information, informing whether the influence of these predictors for each local target predictand is locally or spatially extended. This allows the identification of regions of the target predictand where the relevant spatial structure of the predictors may not be physically plausible, thus indicating potential problems in the models.

\begin{figure}
    \centering
    \includegraphics[width=\linewidth]{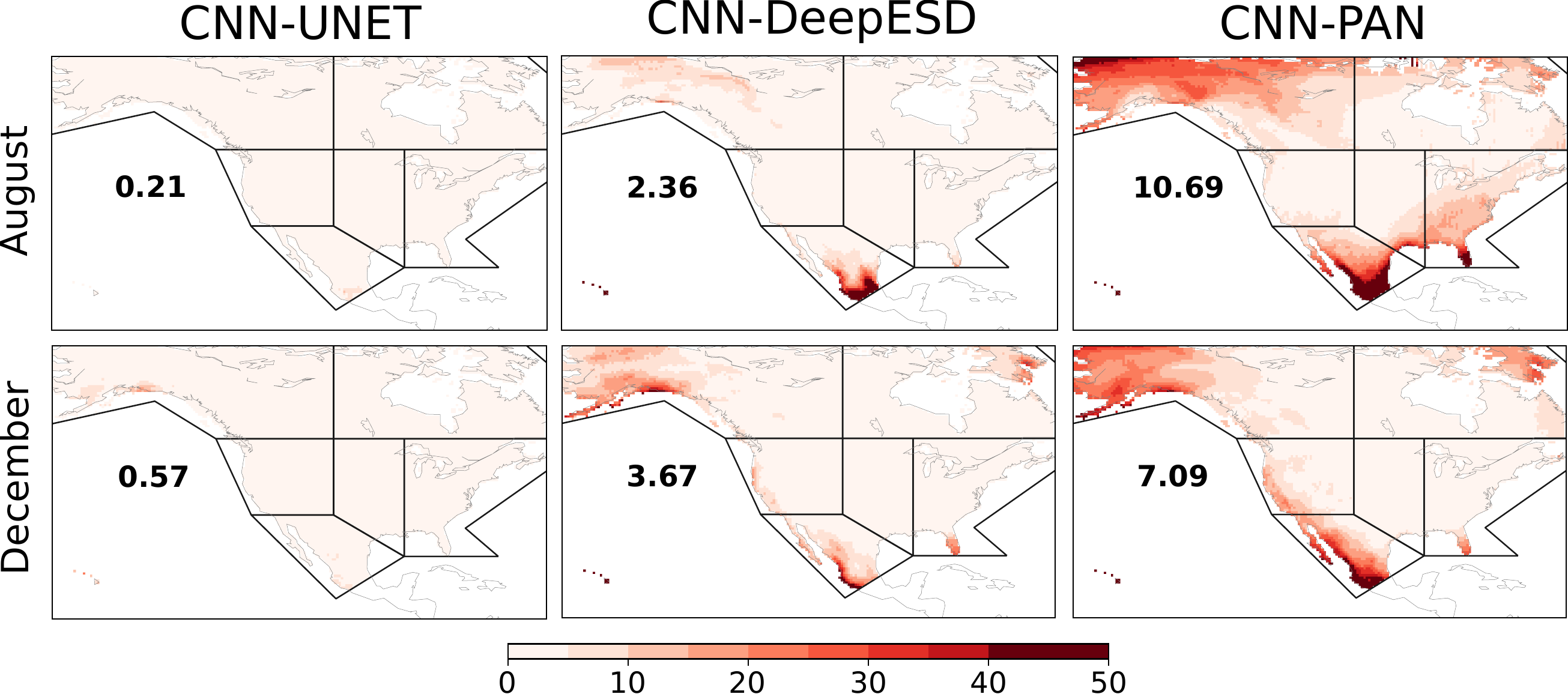}
    \caption{Saliency Dispersion Maps (SDM) for temperature at 1000 hPa for August and December  (in rows) for each of the three intercompared models (in columns).}\label{fig:7}
\end{figure}

Figure \ref{fig:7} shows the SDM results for temperature at 1000 hPa (the most influential variable, from previous results). This variable is shown for each of the intercompared models (in columns) for August and December (in rows). Low values in this metric for a specific gridpoint indicate that the corresponding saliency map for that target predictand is based on predictors localized over the area, while high values indicate higher spatial extent (note that very high results could indicate physically unplausible conditions). For CNN-UNET we can observe how this metric takes values close to zero over the entire North American region (for both August and December), confirming the high degree of locality of this model (as already seen in Figure \ref{fig:6}). For CNN-DeepESD we observe higher values in northern and southern regions, particularly for August in the southern region. This could indicate unplausible wide areas of influence underpinning the model predictions for this region and could explain the large deviation of the climate change signal for the warm season for this region (see Figure \ref{fig:5}). This is even more pronounced and spatially extended for the CNN-PAN model. 

\begin{figure}
    \centering
    \includegraphics[width=\linewidth]{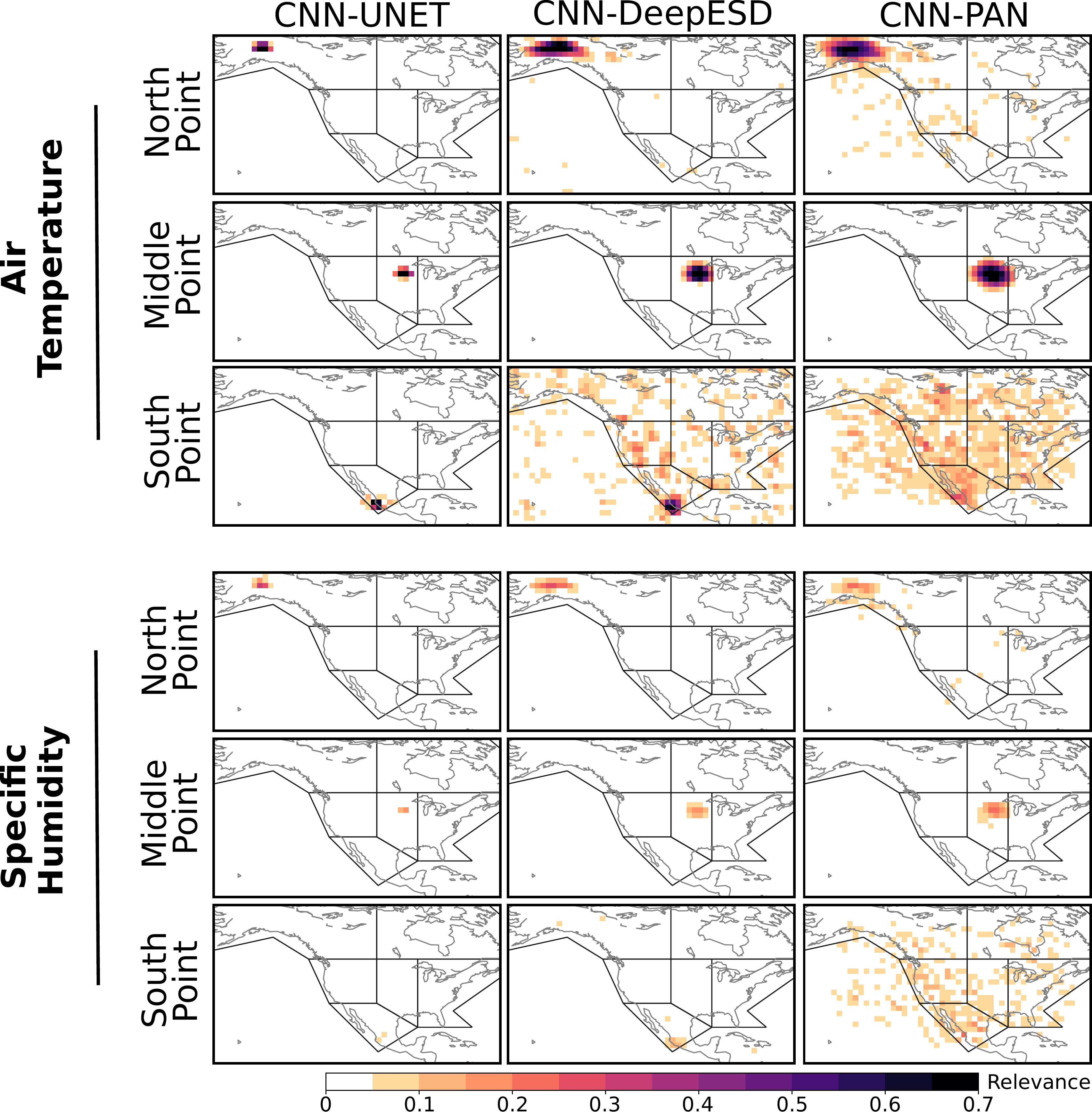}
    \caption{Saliency maps for 1000 hPa temperature and specific humidity for August (top) and December (bottom) for the test period.}\label{fig:8}
\end{figure}

In order to provide a better understanding of this potential problem, 
Figure \ref{fig:8} shows the saliency maps of the different CNN models (in columns) for three illustrative predictand locations representing different latitudes (north, central and south), in rows. For simplicity, we plot the variables air temperature and specific humidity at 1000 hPa. The former corresponds to the most influential variable and the latter is selected to illustrate the behaviour of the other predictors. For the northern point (first and fourth rows, for temperature and humidity, respectively) we see how all models learn a similar localized pattern. Among them, CNN-UNET shows a narrower area of local influence. Furthermore, for CNN-DeepESD and CNN-PAN it can be seen that there are relevant grid points outside the local area, especially for PAN. This non-locality is captured in the SDM metric shown in Figure \ref{fig:7}, where we see high values for the NWN region for these two models, especially for CNN-PAN.  For the central point (second and fifth rows) all models show local areas of influence, with CNN-UNET showing a narrower pattern than CNN-DeepESD and CNN-PAN.  Finally, the southern point exhibit noisy saliency values over the whole area, indicating some structural problem in the CNN-DeepESD and CNN-PAN models, particularly for CNN-PAN. This explains  the lack of extrapolation capability found for these models in this region, producing implausible results for the climate change signals; on the other hand CNN-UNET maintains the same local pattern across the three points indicating no conflict in the learning process.

\section{Conclusions}\label{sec4}
Recently, several studies have shown that deep learning methods (in particular Convolution Neural Network, CNNs) are promising for downscaling applications. These methods have been assessed using standard validation approaches, but they still lack confidence for climate change applications due to lack of explainability. In this work we intercompare several CNN models in the region of North America and introduce several  XAI techniques to analyze the internal behaviour and to inform the evaluation of these methods, revealing potential problem due to structural deficiencies. As a first step we evaluate CNN-UNET, CNN-DeepESD and CNN-PAN considering reanalysis predictors and performing cross-validation. Based on several metrics widely used to evaluate the performance of these models we do not observe clear differences between them. When evaluating these models in the climate model space (GCM predictors) we find discrepancies in the projections of some CNN models (in particular CNN-DeepESD and CNN-PAN), projecting higher temperatures for the southern regions (especially NCA) in comparison with the GCM results, which are used as a pseudo-reality.

In order to provide some insight into this discrepancies we analyze the interval behaviour of the CNN models. We propose and apply several metrics based on XAI techniques. In particular we use saliency maps to  assess the influence of the different predictors in the model's results. We found that  air temperature at 1000 hPa is the most relevant variable for the three models, in agreement with previous studies \cite{maraun_statistical_2018}. However, CNN-DeepESD and CNN-PAN models do not correctly learn the local relationship between predictors and predictand in the southern regions (in particular in NCA), leading to deviations in the downscaled temperature of future scenarios. However, CNN-UNET is able to learn the correct local relationship independently of the region, showing a more robust learning of the concept of locality. These differences are caused by the different topologies of the CNNs. CNN-UNET is a fully convolutional model incorporating the a priori assumption of locality, which ease learning highly local downscaling patterns in the predictors space. However, CNN-DeepESD and CNN-PAN topologies include one and two dense layers at the end of the network, respectively. This facilitates learning a wider spectrum of possible functions and regions of influence, which introduces noise and spurious relationships as seen in the XAI metrics studied in this work. This fact is even more remarkable in the case of CNN-PAN, where two dense layers make it even more difficult to learn the concept of locality. 

The spurious non-local relationship learned by CNN-DeepESD and CNN-PAN causes stronger negative effects in warmer months such as August, as oppose to colder ones like December. This fact aligns with the observed deviations in future periods, which are more pronounced for high percentiles and warm months. This problem of non-locality for CNN-DeepESD and CNN-PAN in the southern region is exacerbated by the extreme latitudinal gradient of North America. Prediction errors in the southern regions with low variability (e.g. NCA) and tropical climate, contribute less to the overall error. Therefore, learning  spurious relationship in these regions with dense layers are not penalized with a large error  and, therefore, these low-variability regions are less optimized in the learning process. 

The results of this work show how the use of XAI techniques allow expanding the toolbox of standard evaluation techniques for assessing downscaling methods based on deep learning. In addition, these techniques help us to understand the internal behavior of the  models and to account for biases and model deficiencies. This is especially relevant when working on large regions where climate variability may induce biases in the learned models. We encourage researchers to incorporate these types of techniques into their evaluation frameworks, especially when working with large regions or under climate change conditions, where a proper understanding of the behaviour of CNN models can help to avoid biases and artifacts in the results.

\section{Data and code availability statement}
All the data required to reproduce the experiments of this work are publicly available. ERA-Interim data can be downloaded from the ECMWF website (\url{https://www.ecmwf.int}) and EC-Earth data from the Earth System Grid Federation (ESGF) portal (\url{https://esgf-data.dkrz.de}). EWEMBI dataset is available at ISIMIP (\url{https://www.isimip.org}). 

The code to reproduce these experiments is publicly available in GitHub (\url{https://github.com/jgonzalezab/XAI-metrics-North-America}). To download and process data and train the models we rely on the \textit{climate}4R framework \cite{iturbide_rframework_2019}. Particularly, for the training of CNN models, we rely on \textit{downscaleR.keras} (\url{https://github.com/SantanderMetGroup/downscaleR.keras}), part of \textit{climate}4R, which uses Tensorflow \cite{whitepaper_tensorflow_2015} and Keras \cite{chollet_keras_2015}. IG technique to compute XAI metrics is implemented through the Python open-source library \textit{innvestigate} \cite{alber_innvestigate_2019}. All these experiments are executed on the Spanish National Research Council (CSIC) deep learning computing infrastructure located at the Instituto de F\'{\i}sica de Cantabria IFCA (CSIC - UC). We train the models in nodes equipped with graphical processing units (GPUs), more specifically NVIDIA Tesla V100 GPUs. In this nodes using one GPU, CNN-DeepESD training takes about 2 seconds per epoch, CNN-PAN 3 seconds and CNN-UNET 50 seconds. To ease the reproducibility of these experiments, as well as to simplify their execution on HPC clusters, we provide the required scripts to follow the workflow presented in \cite{gonzalez_container_2022}. Dockerfiles are also available in the GitHub repository.

\acknowledgments
We acknowledge partial funding from projects ATLAS (PID2019-111481RB-I00) funded by MCIN/AEI 10.13039/501100011033. J. Gonz\'alez-Abad would like to acknowledge the support of the funding from the Spanish Agencia Estatal de Investigaci\'on through the Unidad de Excelencia Mar\'{\i}a de Maeztu with reference MDM-2017-0765. Also, J. Ba\~no-Medina acknowledges support from Universidad de Cantabria and Consejería de Universidades, Igualdad, Cultura y Deporte del Gobierno de Cantabria via the ``instrumentaci\'on y ciencia de datos para sondear la naturaleza del universo'' project.


%
\bibliography{agusample.bib} 
%




%
%
%
%
%

\end{document}